\documentclass[reprint,amsmath,amssymb,aps]{revtex4-1}
\usepackage{bm}

\begin{document}

\title{Protection against Cloning for Deep Learning}

\author{Richard Kenway}
 \email{Email address: r.d.kenway@ed.ac.uk}
\affiliation{School of Physics and Astronomy, University of Edinburgh, James Clerk Maxwell Building, Peter Guthrie Tait Road, Edinburgh EH9 3FD, UK}

\date{\today}

\begin{abstract}
The susceptibility of deep learning to adversarial attack can be understood in the framework of the Renormalisation Group (RG) and the vulnerability of a specific network may be diagnosed provided the weights in each layer are known. An adversary with access to the inputs and outputs could train a second network to clone these weights and, having identified a weakness, use them to compute the perturbation of the input data which exploits it. However, the RG framework also provides a means to poison the outputs of the network imperceptibly, without affecting their legitimate use, so as to prevent such cloning of its weights and thereby foil the generation of adversarial data. 
\end{abstract}


\maketitle

\section{\label{sec:introduction} Introduction}

The emerging understanding of deep learning~\cite{HintonOsinderoTeh2006, LeCunBengioHinton2015, Salakhutdinov2015}, through its analogy with the Renormalisation Group (RG)~\cite{MehtaSchwab2014, LinTegmarkRolnick2017, SchwabMehta2016}, reveals how its vulnerability to adversarial attack~\cite{YuanHeZhuBhatLi2018} is directly related to the depth of the layered network~\cite{Kenway2018}. Successive layers may amplify the sensitivity of the network to specific small perturbations in the input data, causing it to misclassify the input. Whether a deep network has this vulnerability depends on the problem which it is trained to solve and the number of layers. The RG provides a method for diagnosing if the vulnerability exists, using only the trained weights and input data~\cite{Kenway2018}. However, the same information can also be used to construct the specific small changes to the input data that will confuse the network~\cite{Kenway2018}. Thus, it becomes important to prevent a would-be attacker from acquiring the weights.

A possible strategy for an attacker with access to the inputs and outputs of the network would be to use them to train a clone of the network. The cloned weights could then be used to expose any vulnerability and to generate adversarial data. This paper shows how the RG can be used to poison imperceptibly the outputs of the original network, so that, if they are used to clone it, the resulting cloned weights will be wrong and the cloned network will misclassify some data. The attacker would be able to discover that the cloning attempt had failed, but would not be able to rectify it.

In the next section, we summarise the RG framework for deep learning in the context of a deep network in which each layer is a Restricted Boltzmann Machine (RBM)~\cite{Salakhutdinov2015} and the network is trained to classify a data set $\{\bf x\}$ in terms of outputs $\{\bf y\}$, whose exact conditional probability distribution is $p(\bf y|\bf x)$. In Section~\ref{sec:RG for Data Generation} we extend this framework to incorporate the layerwise training, which involves a second deep network, built from the same RBMs, that performs data generation according to $p({\bf x}|{\bf y})$. It is the RG analysis of this data-generation network which is used in Section~\ref{sec:Poisoning} to determine how to poison the outputs of the original classification network, so that it cannot be cloned. Finally, in Section~\ref{sec:conclusions}, we discuss how applicable these RG-inspired methods are for safeguarding deep learning.

\section{\label{sec:RG for classification} Renormalisation Group for Classification}

Consider a layered network of $N$ RBMs with input nodes ${\bf h}_0$, $N-1$ layers of hidden nodes ${\bf h}_k$, $k=1,\ldots,N-1$, and output nodes ${\bf h}_N$. For example, the hidden nodes may be binary vectors of the same dimension as the input data $\{\bf x\}$ and the outputs $\{\bf y\}$, and the joint probability distribution for the $k^{\rm th}$ layer is of the form~\cite{Salakhutdinov2015}
\begin{eqnarray}
t_k({\bf h}_k, {\bf h}_{k-1}) &=& \frac{{\rm e}^{{\bf h}_k^{\rm T} {\bf W}_k {\bf h}_{k-1} + {\bf a}_k^{\rm T}{\bf h}_k + {\bf b}_k^{\rm T}{\bf h}_{k-1}}}{z_k},\nonumber\\ 
z_k &=& {\rm Tr}_{{\bf h}_k, {\bf h}_{k-1}}{\rm e}^{{\bf h}_k^{\rm T} {\bf W}_k {\bf h}_{k-1} + {\bf a}_k^{\rm T}{\bf h}_k + {\bf b}_k^{\rm T}{\bf h}_{k-1}},
\label{eq:RBM}
\end{eqnarray}
where ${\bf W}_k$, ${\bf a}_k$ and ${\bf b}_k$ are weights determined by the layerwise training, discussed further in Section~\ref{sec:RG for Data Generation}.

The trained deep RBM network for the classification problem, $p(\bf y|\bf x)$, generates the probability distribution for the output $\bf y$ given input data $\bf x$, $q_N({\bf y}|{\bf x})$, iteratively for $k=1,\ldots,N$ through the RG-like transformation
\begin{eqnarray}
q_k({\bf h}_k|{\bf x}) &=& {\rm Tr}_{{\bf h}_{k-1}}t_k({\bf h}_k|{\bf h}_{k-1})q_{k-1}({\bf h}_{k-1}|{\bf x}),\nonumber\\
q_0({\bf h}_0|{\bf x}) &=& \delta_{{\bf h}_0, {\bf x}},\nonumber\\
q_N({\bf y}|{\bf x}) &\approx& p({\bf y}|{\bf x}),
\label{eq:deepRBMclassification}
\end{eqnarray}
where, according to Bayes Theorem,
\begin{equation}
t_k({\bf h}_k|{\bf h}_{k-1}) = \frac{t_k({\bf h}_k, {\bf h}_{k-1})}{{\rm Tr}_{{\bf h}_{k}}t_k({\bf h}_k,{\bf h}_{k-1})}.
\end{equation}

We define the effective Hamiltonian for the $k^{\rm th}$ layer of this classification network by
\begin{eqnarray}
q_k({\bf h}|{\bf x}) &=& \frac{{\rm e}^{-H^{(k)}_{\bf x}({\bf h})}}{Z^{(k)}_{\bf x}},\nonumber\\
Z^{(k)}_{\bf x} &=& {\rm Tr}_{\bf h} {\rm e}^{-H^{(k)}_{\bf x}({\bf h})}
\label{eq:classification Hamiltonian}
\end{eqnarray}
and parametrise the space of effective Hamiltonians in terms of a complete set of operators $\{O_\alpha({\bf h})\}$, with couplings $\{g_\alpha\}$. Thus, the effective Hamiltonian for layer $k$ is
\begin{equation}
H^{(k)}_{\bf x}({\bf h}) = \sum_\alpha{g^{(k)}_\alpha O_\alpha({\bf h})}
\label{eq:classification couplings}
\end{equation}
and the effect of the transformation in eq.~(\ref{eq:deepRBMclassification}) is to define $\{g^{(k)}_\alpha\}$ solely in terms of $\{g^{(k-1)}_\alpha\}$, for $k=2,\ldots,N$. This generates a flow in the coupling-constant space of the effective Hamiltonians.

The stability matrix for each layer of the network is defined by
\begin{equation}
T^{(k)}_{\alpha\beta} = \frac{\partial g^{(k)}_\alpha}{\partial g^{(k-1)}_\beta}, \hspace{5mm} k=2,\ldots,N
\label{eq:classification stability matrix}
\end{equation}
and it may be estimated using the method in~\cite{Kenway2018}.

The key assumption we make is that the training of the deep RBM network converges so that, for $N$ large enough, the sequence of effective Hamiltonians converges to a fixed point: 
\begin{equation}
H^*_{\bf x}({\bf h}) = \sum_\alpha{g^*_\alpha O_\alpha({\bf h})},
\label{eq:classification fixed point Hamiltonian}
\end{equation}
such that
\begin{equation}
\frac{{\rm e}^{-H^*_{\bf x}({\bf y})}}{Z^*_{\bf x}} = q_N({\bf y}|{\bf x}) \approx p({\bf y}|{\bf x}).
\label{eq:classification fixed point}
\end{equation}
The stability properties of the fixed points determine whether the network becomes sensitive to small changes in the input data as the number of layers is increased~\cite{Kenway2018}.

\section{\label{sec:RG for Data Generation} Renormalisation Group for Data Generation}

Using the individual RBMs, $t_k({\bf h}_k, {\bf h}_{k-1})$, making up the layers in the deep classification network in eq.~(\ref{eq:deepRBMclassification}), we can construct a deep generation network iteratively for $k=N,\ldots,1$ by effectively reversing the transformation in eq.~(\ref{eq:deepRBMclassification}):
\begin{eqnarray}
\tilde{q}_{k-1}({\bf h}_{k-1}|{\bf y}) &=& {\rm Tr}_{{\bf h}_k}t_k({\bf h}_{k-1}|{\bf h}_k)\tilde{q}_{k}({\bf h}_k|{\bf y}),\nonumber\\
\tilde{q}_N({\bf h}_N|{\bf y}) &=& \delta_{{\bf h}_N, {\bf y}},\nonumber\\
\tilde{q}_0({\bf x}|{\bf y}) &\approx& p({\bf x}|{\bf y}),
\label{eq:deepRBMgeneration}
\end{eqnarray}
where, according to Bayes Theorem,
\begin{equation}
t_k({\bf h}_{k-1}|{\bf h}_k) = \frac{t_k({\bf h}_k, {\bf h}_{k-1})}{{\rm Tr}_{{\bf h}_{k-1}}t_k({\bf h}_k,{\bf h}_{k-1})}.
\end{equation}

In analogy with eqs~(\ref{eq:classification Hamiltonian}) - (\ref{eq:classification stability matrix}), we can define a sequence of effective Hamiltonians for data generation as follows.
\begin{eqnarray}
\tilde{q}_k({\bf h}|{\bf y}) &=& \frac{{\rm e}^{-\tilde{H}^{(k)}_{\bf y}({\bf h})}}{\tilde{Z}^{(k)}_{\bf y}},\nonumber\\
\tilde{Z}^{(k)}_{\bf y} &=& {\rm Tr}_{\bf h} {\rm e}^{-\tilde{H}^{(k)}_{\bf y}({\bf h})},
\label{eq:generation Hamiltonian}
\end{eqnarray}
for $k=N-1,\ldots,0$, with couplings $\{\tilde{g}_\alpha\}$ to the operators $\{O_\alpha({\bf h})\}$ defined by
\begin{equation}
\tilde{H}^{(k)}_{\bf y}({\bf h}) = \sum_\alpha{\tilde{g}^{(k)}_\alpha O_\alpha({\bf h})}
\label{eq:generation couplings}
\end{equation}
and stability matrix
\begin{equation}
\tilde{T}^{(k)}_{\alpha\beta} = \frac{\partial \tilde{g}^{(k-1)}_\alpha}{\partial \tilde{g}^{(k)}_\beta}, \hspace{5mm} k=N-1,\ldots,1.
\label{eq:generation stability matrix}
\end{equation}

Layerwise training of the RBMs $t_k({\bf h}_k, {\bf h}_{k-1})$, $k=1,\ldots,N$, may be achieved by minimising the mutual information,
\begin{equation}
D_{\rm KL}({\rm Tr}_{{\bf x},{\bf y}} \tilde{q}_k({\bf h}_k|{\bf y}) q_{k-1}({\bf h}_{k-1}|{\bf x}) p({\bf x},{\bf y})||t_k({\bf h}_k, {\bf h}_{k-1}))
\label{eq:mutual information}
\end{equation}
with respect to the weights in the $k^{\rm th}$ layer, with the weights in the other layers held fixed, and iterating to convergence for all of the layers. Here $D_{\rm KL}$ is the Kullback-Liebler divergence defined for the two probability distributions $p_1$ and $p_2$ by
\begin{equation}
D_{\rm KL}(p_1||p_2) = {\rm Tr}_{\bf x} p_1({\bf x}) \log \left[ \frac{p_1({\bf x})}{p_2({\bf x})} \right].
\label{eq:KL}
\end{equation}
For classification, the joint probability in eq.~(\ref{eq:mutual information}) is
\begin{equation}
p({\bf x}, {\bf y}) = p({\bf y}|{\bf x})p_{\rm training}({\bf x}),
\label{eq:joint probability}
\end{equation}
where $p_{\rm training}({\bf x})$ is the probability distribution for the sample of classified data, $\{{\bf x}\}$, used for training. The joint probability for the data-generation problem is similar to eq.~(\ref{eq:joint probability}) with ${\bf x}$ and ${\bf y}$ interchanged.

The assumption that the network in eq.~(\ref{eq:deepRBMclassification}) is trained to perform the classification problem correctly, ensures that, as $k$ decreases to 0, the sequence of effective Hamiltonians for the data-generation problem in eq.~(\ref{eq:generation Hamiltonian}) also converges to a fixed point
\begin{equation}
\tilde{H}^*_{\bf y}({\bf h}) = \sum_\alpha{\tilde{g}^*_\alpha O_\alpha({\bf h})},
\label{eq:generation fixed point Hamiltonian}
\end{equation}
such that
\begin{equation}
\frac{{\rm e}^{-\tilde{H}^*_{\bf y}({\bf x})}}{\tilde{Z}^*_{\bf y}} = \tilde{q}_0({\bf x}|{\bf y}) \approx p({\bf x}|{\bf y}).
\label{eq:generation fixed point}
\end{equation}

The stability properties of this fixed point can also be analysed using the method in~\cite{Kenway2018}. If the stability matrix in eq.~(\ref{eq:generation stability matrix}) has an eigenvalue bigger than one for small $k$, so that the fixed point has a relevant direction, then data generation is sensitive to small changes in ${\bf y}$ and these may be utilised to poison the outputs of the classification network without affecting their validity for classification.

\section{\label{sec:Poisoning} Poisoning the Network Outputs}

The classification network may be protected from cloning by adding an imperceptibly small perturbation ${\bf \delta y} \neq {\bf 0}$ to each output ${\bf y}$, which excites an unstable direction of the corresponding data-generation fixed point if it is used for data generation, as it would be in training another network, i.e.,
\begin{equation}
\tilde{q}_0({\bf x}|{\bf y} + {\bf \delta y}) \neq \tilde{q}_0({\bf x}|{\bf y}).
\label{eq:poison}
\end{equation}
The deeper the network, the smaller ${\bf \delta y}$ can be to produce a significant discrepancy in eq.~(\ref{eq:poison}). This renders the mutual information in eq.~(\ref{eq:mutual information}) incorrect for small $k$, so that the resulting cloned weights are wrong and, if ${\bf \delta y}$ is small enough, does not affect the use of ${\bf y} + {\bf \delta y}$ as the classifier.

The perturbation ${\bf \delta y}$ which is the strongest poison, i.e., produces the largest effect for a given small admixture, is proportional to the eigenvector of the Fisher Information Matrix (FIM)~\cite{RajuMachtaSethna2017} for $\tilde{q}_0({\bf x}|{\bf y})$ with the largest eigenvalue~\cite{Kenway2018}. This FIM is given by
\begin{eqnarray}
\tilde{F}^{\scriptscriptstyle (0)}_{ij} &=& \left. \frac{\partial^2}{\partial y'_i \partial y'_j} D_{\rm KL}\left( \tilde{q}_0({\bf x}|{\bf y})||\tilde{q}_0({\bf x}|{\bf y}') \right) \right|_{{\bf y}' = {\bf y}} \nonumber\\
&=& \left\langle \frac{\partial \tilde{H}^{\scriptscriptstyle (0)}_{\bf y}}{\partial y_i} \frac{\partial \tilde{H}^{\scriptscriptstyle (0)}_{\bf y}}{\partial y_j} \right\rangle^{\scriptscriptstyle (0)}
 - \left\langle \frac{\partial \tilde{H}^{\scriptscriptstyle (0)}_{\bf y}}{\partial y_i} \right\rangle^{\scriptscriptstyle (0)} \left\langle \frac{\partial \tilde{H}^{\scriptscriptstyle (0)}_{\bf y}}{\partial y_j} \right\rangle^{\scriptscriptstyle (0)} \nonumber\\
&=& \sum_{\alpha\alpha'}{\frac{\partial \tilde{g}^{\scriptscriptstyle (0)}_\alpha}{\partial y_i} \frac{\partial \tilde{g}^{\scriptscriptstyle (0)}_{\alpha'}}{\partial y_j} \left[ \langle O_\alpha O_{\alpha'}\rangle^{\scriptscriptstyle (0)} - \langle O_\alpha \rangle^{\scriptscriptstyle (0)} \langle O_{\alpha'}\rangle^{\scriptscriptstyle (0)} \right]} \nonumber\\
&&
\label{eq:FIM}
\end{eqnarray}
where $\langle O_\gamma\rangle^{(k)}$ is the expectation value of the operator $O_\gamma$ with respect to the effective Hamiltonian in layer $k$, eq.~(\ref{eq:generation Hamiltonian}), i.e.,
\begin{eqnarray}
\langle O_\gamma\rangle^{(k)} &=& {\rm Tr}_{\bf h} O_\gamma({\bf h}) \tilde{q}_k({\bf h}|{\bf y}) \nonumber\\
                              &=& \frac{{\rm Tr}_{\bf h} O_\gamma({\bf h}){\rm e}^{-\tilde{H}^{(k)}_{\bf y}({\bf h})}}{\tilde{Z}^{(k)}_{\bf y}}.
\label{eq:expectation value}
\end{eqnarray}

Having chosen a subspace of operators, $\{O_\alpha({\bf h})\}$, in which to express the effective Hamiltonians in each layer of the network (via eq.~(\ref{eq:generation couplings})), such that sufficiently precise estimates of the stability matrix associated with each layer, $\tilde{T}^{(k)}_{\alpha\beta}$, in eq.~(\ref{eq:generation stability matrix}) are obtained by the method in~\cite{Kenway2018}, we can compute the FIM using the chain rule:
\begin{equation}
\frac{\partial \tilde{g}^{(0)}_\alpha}{\partial y_i} = \sum_\beta{\left[ \tilde{T}^{(1)} \ldots \tilde{T}^{(N-1)} \right]_{\alpha\beta} \frac{\partial \tilde{g}^{(N-1)}_\beta}{\partial y_i}}.
\label{eq:product of Ts}
\end{equation}
Using eqs~(\ref{eq:deepRBMgeneration}), (\ref{eq:generation Hamiltonian}) and (\ref{eq:generation couplings}), the last derivative may be computed from
\begin{equation}
\frac{{\rm e}^{-\tilde{H}^{(N-1)}_{\bf y}({\bf h}_{N-1})}}{\tilde{Z}^{(N-1)}_{\bf y}} = t_N({\bf h}_{N-1}|{\bf y}),
\label{eq:last layer}
\end{equation}
and the explicit form of the RBM for the $N^{\rm th}$ layer, e.g., using eq.~(\ref{eq:RBM}) to express ${\bf y}^{\rm T} {\bf W}_N {\bf h} + {\bf b}_N^{\rm T} {\bf h}$ in terms of $\{O_\alpha({\bf h})\}$ to determine $\tilde{g}^{(N-1)}_\beta ({\bf y})$.

Thus, the same method used to attack a deep network with poisoned data~\cite{Kenway2018} may be used by the network to protect itself from cloning, provided some of the data-generation fixed points have unstable directions.
\vspace{5mm}

\section{\label{sec:conclusions} Conclusions}

In this paper we have shown how the increasing vulnerability of a deep-learning network to adversarial attack with increasing depth may also provide a means of defence. This applies to a class of deep networks each of which may be represented by layers of RBMs, trained individually so that the network solves a classification problem. The training utilises the same RBMs to construct a data-generation network, with the classifiers as inputs, and it is the potential sensitivity of this network to small changes in the classifiers that turns data poisoning into a means of defence. 

Both the vulnerability to attack and the effectiveness of the defence are proportional to the depth of the network, and are reliant on the corresponding RG fixed points (for classification, or data generation) having a relevant, i.e., unstable direction in the space of operators used to define effective Hamiltonians for each layer. 

If the classification fixed points have no relevant directions, then the network is not susceptible to adversarial attack and no defence is required. This may be ascertained using the method in~\cite{Kenway2018}. If the classification fixed point has a relevant direction, there is no guarantee that the associated data-generation fixed point also has a relevant direction, although this may be determined by the same method, which then also determines the most effective way to poison the classification outputs without affecting their validity as classifiers. 

Use of the poisoned outputs to train a clone of the classification network would cause the cloned network to classify some data incorrectly. While this would be readily apparent to a would-be attacker, it could not be rectified. Unless it can be shown that instability of the classification fixed point implies instability of the associated data-generation fixed point, this protection against cloning is not always available. However, the method in~\cite{Kenway2018} provides a means of checking both the inherent vulnerability of a given network and whether poisoning of its outputs is an effective defence, so that the safety of the network can be established.

\end{document}